\documentclass[10pt,letterpaper]{article}
\usepackage{graphicx}
\usepackage{amsmath}
\usepackage{amssymb}
\usepackage{booktabs}
\usepackage{caption}
\usepackage{subcaption}
\usepackage{float}

\usepackage{etoolbox}
\makeatletter
\newcommand{\paperaward}{%
  \vspace{1em}%
  \noindent\parbox{\textwidth}{%
    \centering
    \textbf{Best Paper Award -- Accepted at the International Conference on AI for the Oceans (ICAIO) 2025.}\\
    \textit{To appear in Springer Lecture Notes in Networks and Systems (LNNS). This is the pre-publication version.}%
  }%
}

\apptocmd{\@maketitle}{\paperaward}{}{}
\makeatother

\title{Advancing Welding Defect Detection in Maritime Operations via Adapt-WeldNet and Defect Detection Interpretability Analysis}

\author{
Kamal Basha S and Athira Nambiar\\
Department of Computational Intelligence,\\
Faculty of Engineering and Technology,\\
SRM Institute of Science and Technology,\\
Kattankulathur, Tamil Nadu, 603203, India\\
{\tt\small c58527@srmist.edu.in, athiram@srmist.edu.in}
}
\date{}  

\usepackage[pagebackref]{hyperref}
\usepackage[capitalize]{cleveref}
\hypersetup{
    pdfauthor={Kamal Basha S, Athira Nambiar},
    pdftitle={Advancing Welding Defect Detection in Maritime Operations via Adapt-WeldNet and Defect Detection Interpretability Analysis}
}

\begin{document}
\maketitle


\begin{abstract}
Weld defect detection is crucial for ensuring the safety and reliability of piping systems in the oil and gas industry, especially in challenging marine and offshore environments. Traditional non-destructive testing (NDT) methods often fail to detect subtle or internal defects, leading to potential failures and costly downtime. Furthermore, existing neural network-based approaches for defect classification frequently rely on arbitrarily selected pretrained architectures and lack interpretability, raising safety concerns for deployment. To address these challenges, this paper introduces ``Adapt-WeldNet", an adaptive framework for welding defect detection that systematically evaluates various pre-trained architectures, transfer learning strategies, and adaptive optimizers to identify the best-performing model and hyperparameters, optimizing defect detection and providing actionable insights. Additionally, a novel Defect Detection Interpretability Analysis (DDIA) framework is proposed to enhance system transparency. DDIA employs Explainable AI (XAI) techniques, such as Grad-CAM and LIME, alongside domain-specific evaluations validated by certified ASNT NDE Level II professionals. Incorporating a Human-in-the-Loop (HITL) approach and aligning with the principles of Trustworthy AI, DDIA ensures the reliability, fairness, and accountability of the defect detection system, fostering confidence in automated decisions through expert validation. By improving both performance and interpretability, this work enhances trust, safety, and reliability in welding defect detection systems, supporting critical operations in offshore and marine environments.


\end{abstract}

\section{Introduction}
\label{sec:intro}
\vspace{-0.2cm}
Welding is a fabrication process in which two or more materials, typically metals, are joined together by causing coalescence through the application of heat, pressure, or both. It plays a pivotal role in the construction and maintenance of maritime offshore structures, including oil rigs, pipelines, and support vessels, due to its efficiency and versatility~\cite{jeffus1999welding}. In offshore applications, precise welding is essential to ensure the structural integrity of components that must withstand harsh marine environments, dynamic loads, and corrosive conditions~\cite{manzano2024multiphase}. 

Despite its critical importance, welding in maritime offshore structures is susceptible to various defects such as cracks, porosity, and lack of penetration, which can severely compromise safety and performance~\cite{standard2004welding}. Non-Destructive Testing (NDT) techniques~\cite{gupta2022advances}, including ultrasonic, radiographic, and magnetic particle testing, are essential for identifying these flaws and ensuring compliance with maritime safety standards. However, traditional NDT methods may struggle to detect subtle or internal defects effectively, especially under challenging offshore conditions. In recent years, advanced AI-driven techniques have emerged as transformative tools, offering improved defect detection accuracy, automating inspection processes, and enhancing maintenance strategies, all of which contribute to the reliability and longevity of offshore structures~\cite{silva2021x}. However, challenges remain in applying these advanced technologies to welding defect detection, particularly in offshore environments where domain-specific issues such as harsh marine conditions complicate matters. Pre-trained models, while effective in some general scenarios, often underperform in these contexts due to a lack of optimization for domain-specific parameters.

To overcome these limitations, we propose a novel framework ``\textbf{Adapt-WeldNet}", that adaptively optimizes the model through adaptive transfer Learning, adaptive model selection, and adaptive optimizer and hyperparameters. This approach is specifically designed for welding defect detection in offshore environments. By dynamically selecting the best parameters and model configurations, Adapt-WeldNet aims to deliver the most effective trial for developing a classifier. This ensures improved performance in detecting and addressing welding flaws, tailored to the unique and challenging conditions of offshore structures.

Despite the promising capabilities of AI in defect detection, yet another critical limitation is the lack of interpretability in black-box models. The inability to understand how models arrive at their decisions undermines trust and hinders their adoption, especially in high-stakes applications like welding defect detection in offshore environments. To this end, Explainable AI (XAI) has emerged as an efficient solution towards providing transparency, accountability, and trust in machine learning models by making their decision-making processes more understandable and interpretable to humans. Recent works have proposed various evaluation metrics for XAI methods, including reference-based metrics such as Pearson correlation coefficient (PCC) and similarity (SIM), which compare explanation maps with human-interpreted maps or pixel importance in input images~\cite{bourroux:hal-03689004}. The Defense Advanced Research Projects Agency (DARPA)~\cite{gunning2019darpa} also highlights the importance of XAI in critical applications. In these scenarios, where safety and reliability are paramount, XAI plays a key role in building trust and ensuring the adoption of AI-driven defect detection systems.

In welding defect detection, the application of XAI is still underexplored. In this regard, we propose the \textbf{Defect Detection Interpretability Analysis (DDIA) framework}, which incorporates Human-in-the-Loop and Trustworthy AI principles. This framework allows domain experts in welding to provide human interpretations, ensuring the transparency and reliability of AI-driven defect detection. Furthermore, DDIA includes quantitative metrics for interpretability in defect localization, enabling an objective evaluation of defect detection models. As part of this framework, we propose a novel \textbf{recall-based evaluation metric} for the interpretability of XAI techniques in defect localization, marking this as the first domain-specific evaluation of XAI in welding defect detection. The key contributions of this paper include:
\begin{itemize}
\item A novel \textbf{"Adapt-WeldNet"}, an adaptive AI framework for optimizing the model selection and parameter tuning in welding defect detection.
\item The \textbf{Defect Detection Interpretability Analysis (DDIA) }framework, a novel approach for evaluating the performance of different XAI methods through domain expert analysis, one of its first kind.
\item Proposal of a novel \textbf{`recall-based evaluation metric' }for the interpretability of XAI techniques for defect localization.
\item Extensive quantitative and qualitative analysis on a real-world welding dataset.
\end{itemize}


\section{Related works}
\vspace{-0.2cm}

Traditional welding defect detection methods, like visual inspection and NDT, face accuracy limitations. A machine vision algorithm~\cite{sun2019effective} has achieved 95\% accuracy, reaching up to 99\% in real-world applications.
Recent advancements in weld defect detection have explored a variety of deep learning models, including both pre-trained models and novel architectures. \cite{VASAN2024108961} presents an ensemble-based deep learning model for welding defect detection and classification, achieving 93.12\% accuracy in Non-Destructive Testing (NDT) of submerged arc welds. Pre-trained models such as AlexNet, VGG-16, VGG-19, ResNet50, and GoogLeNet have demonstrated excellent performance, particularly with the application of data augmentation techniques on datasets like GDXray~\cite{ajmi2020deep}. Fine-tuning strategies using VGG16 and ResNet50 on augmented X-ray datasets have also yielded effective results, with VGG16 achieving 90\% accuracy despite data imbalance~\cite{kumar2023semi}. In specific applications like TIG welding defect classification, models such as VGG16, VGG19, ResNet50, InceptionV3, and MobileNetV2 have been successfully applied with various optimizers, achieving high accuracy~\cite{sekhar2022intelligent}.

Transfer learning has played a key role in improving defect detection across different welding techniques. For instance, transfer learning with GoogLeNet and Multi-Layer Perceptron (MLP) has resulted in a 96.99\% classification accuracy for spot-welding evaluations~\cite{yang2018evaluation}. Moreover, a deep transfer learning model has been successfully applied to aeronautics composite materials, achieving 96\% accuracy and demonstrating robustness even with limited labeled data~\cite{gong2020deep}.

In addition to traditional models, recent work has explored more advanced architectures like the Vision Transformer (ViT), which has been incorporated into a Deep Feature Extraction Module (DFEM) for enhanced X-ray weld defect detection~\cite{zhang2024research}. Similarly, VAE-DCGAN—a combination of a Variational Autoencoder (VAE) with a Deep Convolutional GAN (DCGAN)—has been employed for improved feature representation in weld joint defects, leading to better defect detection performance~\cite{geng2024gan}. Furthermore, Faster RCNN-based models have also been explored for their potential in real-time weld defect detection applications~\cite{ajmi2024advanced}. Finally, real-time applications, such as WelDeNet, a deep CNN with 14 convolutional layers, have achieved remarkable accuracy (99.5\%) in welding defect classification, demonstrating their applicability in practical, time-sensitive environments~\cite{perri2023welding}.

In contrast to the aforementioned works, ``Adapt-WeldNet" offers an optimal solution by leveraging adaptive transfer learning, adaptive models, and adaptive optimizers. Similarly, the Defect Detection Interpretability Analysis (DDIA) framework goes beyond applying XAI by integrating domain experts into the loop, fostering a trustworthy AI system tailored for the maritime offshore industrial environment. 


\vspace{-0.5cm}
\section{Methodology}
\label{sec:methodology}
\begin{figure}[h]
    \centering
    \includegraphics[width=1.15\columnwidth]{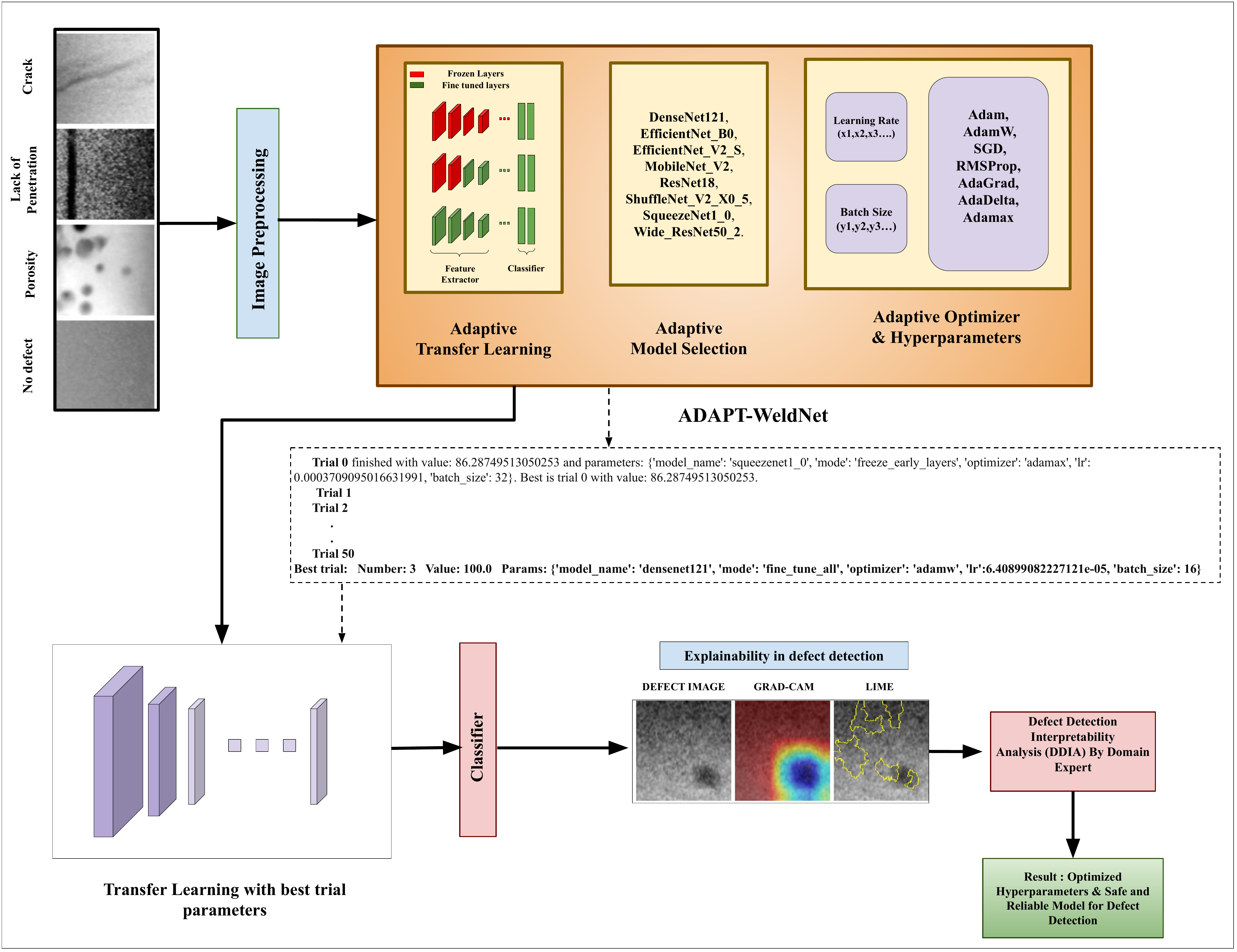}
    \caption{Proposed \textbf{Adapt-WeldNet} framework for welding defect detection, integrating adaptive models, adaptive transfer learning, adaptive optimizers, and hyperparameter tuning, alongside the DDIA framework with Explainable AI (XAI) for enhanced interpretability and performance.}
    \label{fig:architecture_diagram}
\end{figure}

\vspace{-0.2cm}
In this section, we present the comprehensive methodology adopted in this study. Figure~\ref{fig:architecture_diagram} illustrates the architectural workflow of the proposed Adapt-WeldNet framework. Initially, images representing 4 types for images i.e. crack (C), lack of penetration (LP), porosity (P), and no defect are fed into an image preprocessing pipeline. After preprocessing, the Adapt-WeldNet framework optimizes model performance under varying conditions, incorporating adaptive transfer learning, adaptive model selection, and adaptive optimizer and hyperparameter tuning, as discussed in Section~\ref{Adapt-weldnet}. The best-performing classifier, identified through frm Adapt-WeldNet framework, is further developed and evaluated using explainable AI (XAI) techniques, detailed in Section~\ref{Explainability_Techniques}. Further advancing this, the proposed Defect Detection Interpretability Analysis (DDIA) framework enhances the evaluation of XAI by incorporating assessments from welding experts, thereby improving the interpretability and reliability of the results. The DDIA framework is discussed in detail in Section~\ref{sec:ddia_framework}.
After receiving expert validation, the model is deployed for safe and transparent defect detection.
    

\vspace{-0.3cm}

\subsection{Adapt-WeldNet framework}
\label{Adapt-weldnet}
\vspace{-0.2cm}

Most existing works on weld defect detection relies on using random architectures with arbitrary parameters, which may not be well-suited for this critical domain. This lack of systematic optimization can lead to suboptimal results, raising concerns about reliability and trustworthiness during deployment, especially in critical maritime offshore structures. To address this, the Adapt-WeldNet framework is designed to optimize model selection and parameter tuning. By systematically identifying the best-performing architectures and parameters, this approach ensures greater transparency and trustworthy AI, making it more suitable for high-stakes applications.

Initially, the input defect images are fed into a preprocessing pipeline, where operations such as image resizing and normalization are performed. The preprocessed images are then passed to the Adapt-WeldNet framework, which employs adaptive model selection. This process involves evaluating eight neural network architectures pre-trained on ImageNet, utilizing three adaptive transfer learning strategies. Additionally, the framework explores seven different optimizers, along with varying learning rates and batch sizes, to identify the best model tailored for this specific domain. The optimization process is guided by the Optuna library~\cite{akiba2019optuna} , and further details are provided below.

\begin{itemize}
    \item \textbf{Adaptive Model Selection:} The model name is chosen from the following options: ResNet18, DenseNet121, EfficientNet-B0, EfficientNet-V2-S, MobileNet-V2, Wide ResNet50-2, ShuffleNet V2 X0.5, and SqueezeNet1.0.
    \item \textbf{Adaptive Transfer Learning:} Transfer learning \cite{yosinski2014transfer} involves leveraging a pre-trained model on a similar task and adapting it to a new task, often by fine-tuning certain layers. The model training mode is selected from three options: 
    \begin{itemize}
        \item \textit{Freeze Early Layers:} In this mode, the early layers of the pre-trained model are frozen to retain learned features, while the later layers, including the classifier, are fine-tuned to adapt to the new dataset.
        \item \textit{Freeze All Layers:} Here, all layers of the pre-trained model except for the final classifier are frozen, and only the classifier is trained to adapt to the new task.
        \item \textit{Fine-tune All:} This approach involves fine-tuning all layers of the pre-trained model to fully adapt it to the new dataset.
    \end{itemize}

    \item \textbf{Adaptive Optimizer and Hyperparameters:} The optimizer is chosen from the following options: Adam~\cite{kingma2014adam}, AdamW~\cite{loshchilov2017decoupled}, SGD\cite{robbins1951stochastic}, RMSprop~\cite{hinton2012lecture}, Adagrad~\cite{duchi2011adaptive}, Adadelta~\cite{zeiler2012adadelta}, and Adamax~\cite{kingma2014adam}. These optimizers are designed to improve model convergence during training. Additionally, the \textbf{learning rate} is optimized from a log-uniform distribution between \( 1 \times 10^{-5} \) and \( 1 \times 10^{-2} \) and the \textbf{batch size} is chosen from the options: 16, 32, and 64.
\end{itemize}

The Adapt-WeldNet framework systematically selects the optimal combination of parameters through trial and error, ultimately resulting in the best-performing model configuration for weld defect detection. The classifier is developed using the best hyperparameters identified through this process. To further enhance the interpretability and transparency of the model, additional explainable AI (XAI) techniques, such as Grad-CAM and LIME, are applied. 

\vspace{-0.2cm}

\subsection{Explainability Techniques}
\label{Explainability_Techniques}
\vspace{-0.1cm}
Grad-CAM and LIME are employed to provide insights into the classifier's decision-making process, offering a deeper understanding of its predictions. Below are the mathematical formulations and implementation details of these techniques.

\subsubsection{Grad-CAM (Gradient-weighted Class Activation Mapping)}

Grad-CAM~\cite{Selvaraju_2017_ICCV} generates a heatmap by computing the weighted sum of the feature maps from the last convolutional layer, where the weights are determined by the gradients of the output class score \( y^c \) with respect to these feature maps:

\begin{equation}
L^c = \text{ReLU}\left( \sum_k \left( \frac{1}{Z} \sum_{i,j} \frac{\partial y^c}{\partial A_{ij}^k} \right) A^k \right)
\end{equation}

In this equation, \( L^c \) represents the Grad-CAM heatmap for class \( c \), highlighting the regions of the image that influence the model's prediction for that class. The function \( \text{ReLU} \) ensures only positive contributions are considered. The term \( \sum_k \) denotes the summation over all feature maps \( k \). The weights for each feature map are computed by averaging the gradients of the class score \( y^c \) with respect to the feature map activations \( A_{ij}^k \) over all spatial locations \( (i, j) \), normalized by the total number of pixels \( Z \) in the feature map. Finally, each feature map \( A^k \) is weighted by its corresponding importance and summed to produce the heatmap.

\subsubsection{LIME (Local Interpretable Model-agnostic Explanations)}

LIME~\cite{ribeiro2016whyitrustyou} explains predictions by perturbing input data and fitting a simple model to the perturbed samples, minimizing the locality-aware loss:

\begin{equation}
L(f_p, g, \pi_x) = \sum_{z \in Z} \pi_x(z) \left(f_p(z) - g(z)\right)^2
\end{equation}

Where \( f_p(z) \) is the prediction of the complex model, \( g(z) \) is the simpler model, and \( \pi_x(z) = \exp\left(-\frac{D(x, z)^2}{\sigma^2}\right) \) gives higher weight to samples close to \( x \), ensuring a local approximation of the complex model.

\subsection{Defect Detection Interpretability Analysis (DDIA) Framework with Expert Guided Evaluation}
\label{sec:ddia_framework}
\vspace{-0.1cm}
\begin{figure}[h]
    \centering
    \includegraphics[width=\textwidth]{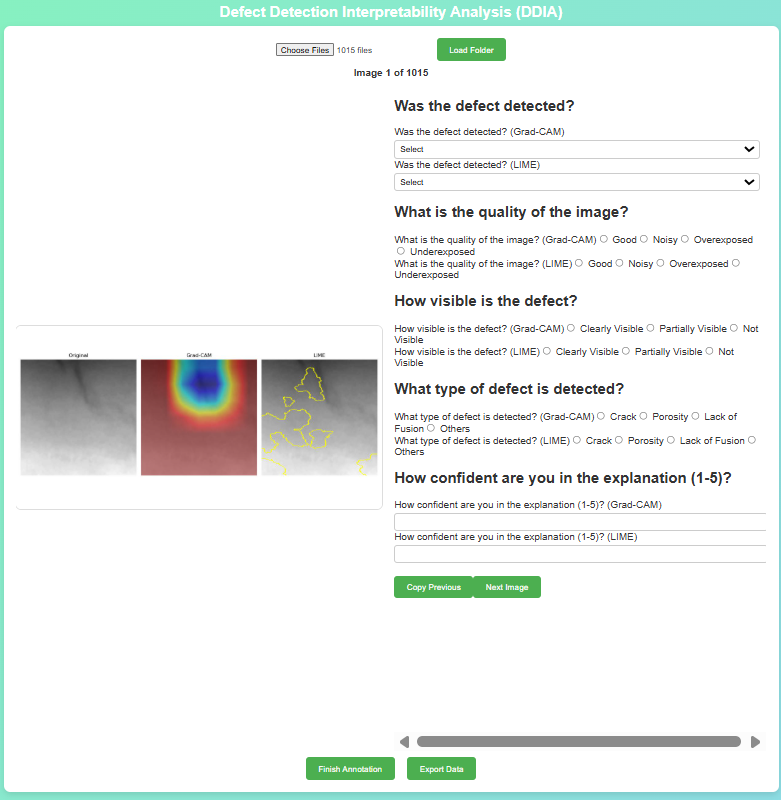}
    \caption{Defect Detection Interpretability Analysis (DDIA) Framework}
    \label{fig:DDIA Framework}
\end{figure}

While explainable AI (XAI) plays a pivotal role in making deep learning models interpretable, the Defect Detection Interpretability Analysis (DDIA) framework advances this by incorporating a human-in-the-loop approach, involving certified domain experts. This ensures a comprehensive evaluation of the AI model, addressing not only interpretability but also reliability and alignment with real-world requirements. In particular, the DDIA framework leverages feedback from specialists, such as ASNT NDE Level II~\cite{asnt_website} auditors, who assess critical factors like detection accuracy, defect visibility, image quality, defect type, and prediction confidence. By analyzing XAI outputs, such as Grad-CAM and LIME explainers, these experts trace the model’s decision-making process and identify areas for improvement, such as refining the dataset or retraining the model to enhance robustness.

The DDIA framework as shown in Figure \ref{fig:DDIA Framework} is designed to tackle real-world challenges encountered in X-ray imaging, such as overexposure, underexposure, noise. It incorporates a structured evaluation mechanism, wherein XAI results are presented to experts under various X-ray film scenarios. This mechanism includes targeted questions aimed at assessing:
The defect detection is evaluated by determining whether the defect was detected (Yes/No for both Grad-CAM and LIME). The image quality is assessed as clear, underexposed, overexposed, or noisy. Defect visibility is rated as clearly visible, partially visible, or not visible for both Grad-CAM and LIME. The defect type is categorized as crack, lack of penetration (LP), or porosity. Lastly, auditors assign confidence scores ranging from 1 to 5 for both Grad-CAM and LIME explanations.

These structured inputs enable a detailed analysis of the XAI results, categorizing X-ray films by quality (e.g., clear, noisy, overexposed, underexposed) and identifying the strengths of each explainer. For instance, the analysis reveals which XAI method, Grad-CAM or LIME, provides higher confidence to auditors and performs better under specific conditions.

By integrating domain expertise into the evaluation process, the DDIA framework fosters trust and ensures safer deployment of AI systems in high-stakes environments like offshore marine structures. This human-in-the-loop approach not only enhances model transparency but also addresses critical safety concerns, making it one of the first frameworks to combine XAI and expert audits in welding defect detection for maritime applications.

\vspace{-0.5cm}

\section{Experimental Setup}
\vspace{-0.3cm}
In this section, we present the experimental setup, including the dataset described in Section~\ref{sec:dataset} and the evaluation metrics outlined in Section~\ref{sec:evaluation_metrics}, which are applied to assess the performance of the Adaptive AI model selection and Explainable AI techniques.
\vspace{-0.4cm}
\subsection{Dataset}
\label{sec:dataset}
\vspace{-0.2cm}
The experiments utilized the RIAWELC dataset~\cite{totino2023riawelc}, that consists of 24,407 X-ray weld images categorized into four defect classes: porosity(P), lack of penetration (LP), crack(C), and no defects(ND). For each experiment, the dataset is divided into training and validation sets, ensuring a balanced representation of each class in both sets. Initially, the class distribution in the training dataset was uneven, with 7008 samples for crack, 3712 samples for low penetration, 5352 samples for no defects, and 5768 samples for porosity. After applying data augmentation techniques, the dataset was balanced, resulting in 7008 samples for each class.
\vspace{-0.3cm}
\subsection{Evaluation Metrics}
\label{sec:evaluation_metrics}
\vspace{-0.2cm}
\subsubsection{Accuracy and Confusion Matrix}
\vspace{-0.2cm}
Accuracy measures the proportion of correctly classified instances, calculated as the ratio of true positives (TP) and true negatives (TN) to the total instances, including false positives (FP) and false negatives (FN). A confusion matrix provides a detailed breakdown of the classification performance by showing TP, TN, FP, and FN, helping to understand the model's errors and successes~\cite{domingos2012few, fawcett2006introduction}.
\vspace{-0.5cm}
\subsubsection{Quantitative Metrics for Interpretability in Defect Localization}
\vspace{-0.2cm}
To assess the interpretability of Grad-CAM in localizing welding defects, we employ a novel recall-based evaluation method. This quantitative metric measures the ability of Grad-CAM to correctly identify defect regions in welding images. The dataset used for this evaluation comprises 1,031 annotated images, including defects such as cracks, porosity, and lack of penetration (LP), with annotations provided by certified domain experts. 

Recall, a key metric for this analysis, is defined as:

\begin{equation}
\text{Recall} = \frac{\sum_i (L_i^c \cdot G_i)}{\sum_i G_i}
\end{equation}

Here, \( L_i^c \) represents the Grad-CAM heatmap for the \(i\)-th image and class \(c\), while \( G_i \) denotes the ground truth defect region for the \(i\)-th image. This formulation ensures that the overlap between Grad-CAM predictions and expert-annotated ground truth regions is accurately measured.

To evaluate overall performance, we compute the Average Recall, defined as the mean of recall values across all images in the dataset:

\begin{equation}
\text{Average Recall} = \frac{1}{N} \sum_{i=1}^{N} \frac{\sum (L_i^c \cdot G_i)}{\sum G_i}
\end{equation}

Where \( N \) represents the total number of images. This metric provides a comprehensive evaluation of Grad-CAM's ability to localize defect regions across a diverse dataset.

Recall measures how much of the defect area the model was able to detect. A recall of 1 (or 100\%) means the model found all the defect pixels in the ground truth. A recall of 0.5 means the model detected only half of the defect pixels. This shows how well Grad-CAM identifies defect regions, with varying recall values reflecting the model's performance across different defects and image conditions.

\vspace{-0.4cm}

\subsection{Implementation details}
\vspace{-0.2cm}
The training procedure involved selecting various models such as DenseNet121, EfficientNet-B0, MobileNet-V2, ResNet18, ShuffleNet-V2 x0.5, SqueezeNet1.0, and WideResNet50-2, each initialized with pre-trained ImageNet weights. The final fully connected layer was adjusted to match the output classes of the dataset. Three training modes were employed: Freeze All Layers (where all layers are frozen and only the classifier head is trained), Freeze Early Layers (where the early layers are frozen and only the deeper layers and classifier head are trained), and Fine-Tuning All Layers (where all model parameters are updated).

For model optimization, we utilized the Adapt-Weld framework to tune hyperparameters, including model architecture, optimizer type, learning rate, and batch size. The function \texttt{train\_and\_evaluate} handles the training and evaluation process. The model, training mode, and dataset are specified, and the Cross-Entropy Loss is used as the loss function. The selected optimizer is applied during training, and early stopping is employed with a patience value to prevent overfitting.

The \texttt{objective} function defines the search space for Optuna, allowing it to suggest the best model architecture, optimizer type, learning rate, and batch size. Specifically, the model architecture (e.g., ResNet18, DenseNet121, EfficientNet-B0, etc.) will be selected from Adapt-Weld framework, the training mode (Freeze All Layers, Freeze Early Layers, or Fine-Tuning All Layers), the optimizer type (Adam, AdamW, SGD, RMSProp, AdaGrad, AdaDelta, or Adamax), the learning rate (ranging from 1e-5 to 1e-2), and the batch size (16, 32, or 64).

The model is loaded according to the trial’s suggestions, and the corresponding training configuration is applied. The training loop runs for 100 epochs, with early stopping if the validation accuracy does not improve for five consecutive epochs. After each epoch, the model’s performance is evaluated on the validation set.

\vspace{-0.5cm}

\section{Experimental Results}
\vspace{-0.2cm}
The experimental results of our proposed Adapt-WeldNet framework is presented in this section. We first discuss the quantitative results in Section \ref{qant_adapt}, followed by qualitative results of Explainable AI (XAI) in Section \ref{xai_qual}. Section \ref{quant_ddia} covers the quantitative analysis of the Defect Detection Interpretability Analysis (DDIA) framework. Lastly, we provide the quantitative metrics for interpretability in defect localization in Section \ref{quant_inter}.

\subsection{Quantitative Results of ADAPT-WeldNet}
\label{qant_adapt}
\subsubsection{Adapt-WeldNet Optimization Results:} 

\vspace{-0.2cm}
\begin{figure}[htbp]
    \centering
    \includegraphics[width=1.2\linewidth]{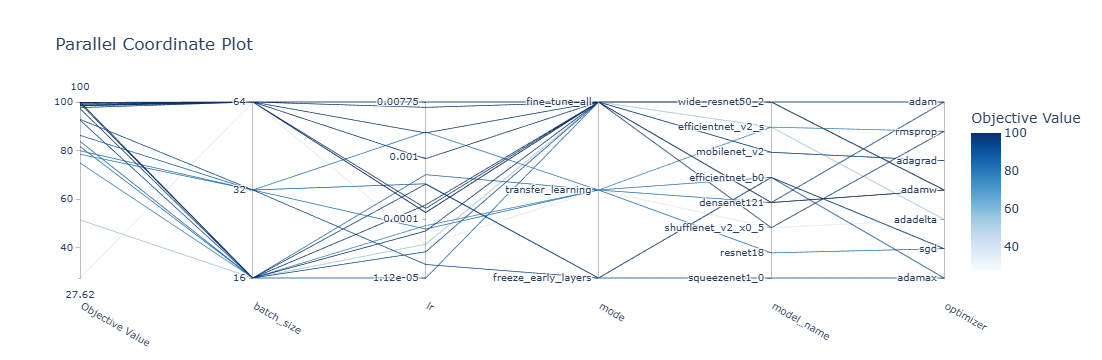}
    \caption{Parallel coordinate plot showing the relationship between hyperparameters and the objective value in the adaptive AI model optimization process. Darker colors indicate higher objective values.(Best viewed in colour)}
    \label{fig:parallel_coordinates}
\end{figure}

The parallel coordinate plot in Figure~\ref{fig:parallel_coordinates} illustrates the impact of various hyperparameters, including batch size, learning rate, mode, model architecture, and optimizer, on the objective value during the optimization of the adaptive AI model for welding defect detection. The results demonstrate that fine-tuning all layers consistently achieves superior performance compared to freezing early layers. Models such as \textit{EfficientNet\_B0}, \textit{EfficientNet\_V2\_S}, and \textit{DenseNet121} exhibit the best performance when combined with optimizers like AdamW or Adam at lower learning rates (e.g., $1 \times 10^{-5}$). Additionally, extreme batch sizes—both small (16) and large (64)—are associated with higher objective values, while medium batch sizes (32) show more variability. These findings underscore the significance of hyperparameter tuning and adaptive AI techniques in optimizing model performance for specific domains like welding defect detection.

\begin{figure}[h!]
    \centering
    \includegraphics[width=1.1\linewidth]{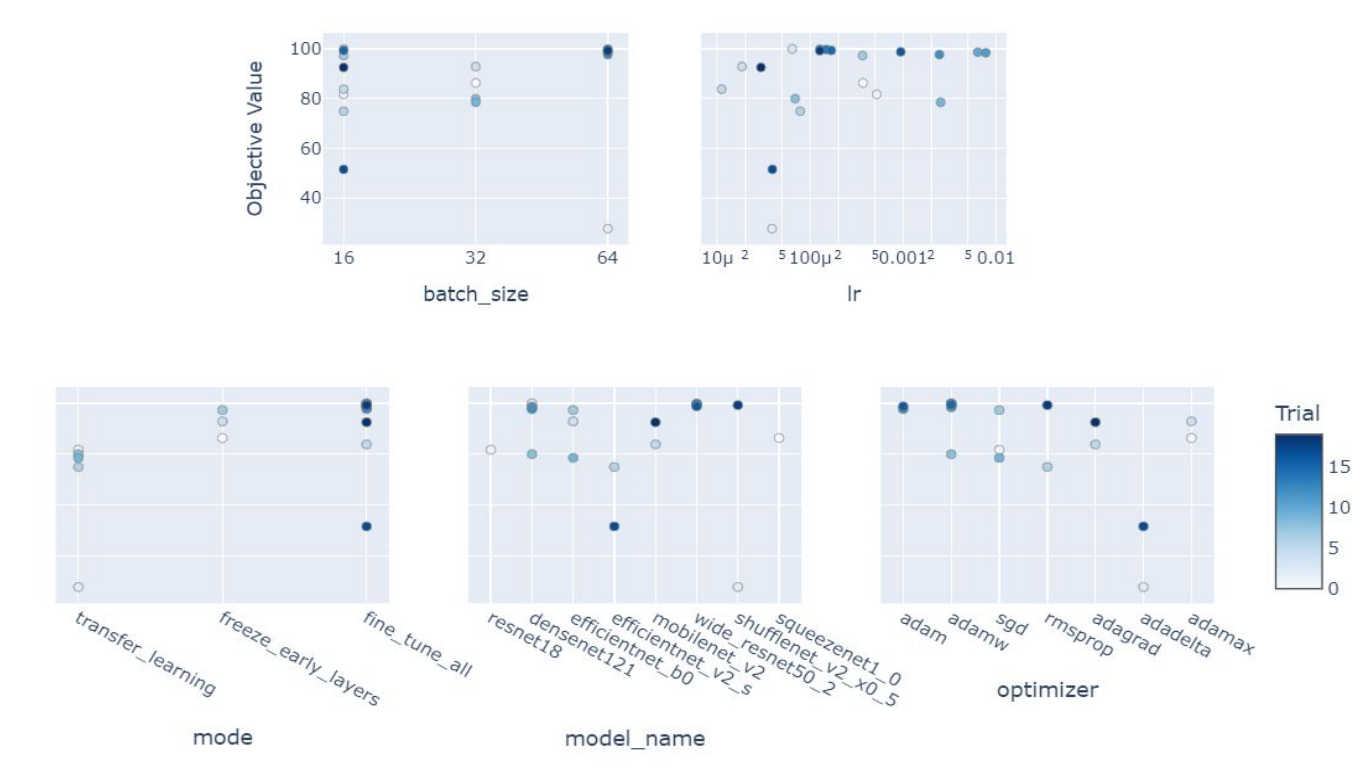}
    \caption{Hyperparameter tuning results of  Adapt-WeldNet for batch size, learning rate, training mode, model architecture, and optimizer.}
    \label{fig:optuna_results}
\end{figure}

Figure~\ref{fig:optuna_results} shows the Adapt-Weld framework optimization results for hyperparameter tuning across various configurations, including batch size, learning rate (lr), training mode, model architectures, and optimizers. The objective value represents the validation accuracy achieved for each trial, with the color intensity indicating the trial number.

Models with a batch size of 16 generally achieved higher objective values compared to larger batch sizes (32 and 64). Learning rates in the range of \(10^{-5}\) to \(10^{-3}\) showed consistent performance, with a few trials achieving near-optimal objective values. The fine-tune-all mode outperformed the other modes (transfer learning and freezing early layers), highlighting the benefit of updating all layers during training. DenseNet121 and WideResNet50-2 consistently showed superior objective values compared to other models such as MobileNet-V2 and ShuffleNet-V2 x0.5. Among the optimizers, Adam and AdamW showed the best performance, whereas SGD and RMSprop exhibited slightly lower objective values in most trials.

These results emphasize the importance of hyperparameter tuning to achieve optimal performance and reveal the sensitivity of performance to specific configurations.
\vspace{-0.2cm}
\vspace{-0.3cm}

\subsubsection{Effect of Parameter Tuning Modes:} 

Figure~\ref{fig:params_mode_boxplot1} presents the comparative performance of three parameter tuning modes: \textit{transfer learning}, \textit{freeze\_early\_layers}, and \textit{fine\_tune\_all}. The box plot highlights that fine-tuning all layers achieves the highest performance consistency, with minimal variation and only a few outliers. Freezing early layers also maintains robust performance, whereas transfer learning in freezing all layers shows comparatively lower values and larger variability.

\subsubsection{Classifier Performance Evaluation from Best Hyperparameters:} 

\begin{figure}[htbp]
    \centering

    \begin{minipage}{0.45\linewidth}
        \centering
        \includegraphics[width=\linewidth]{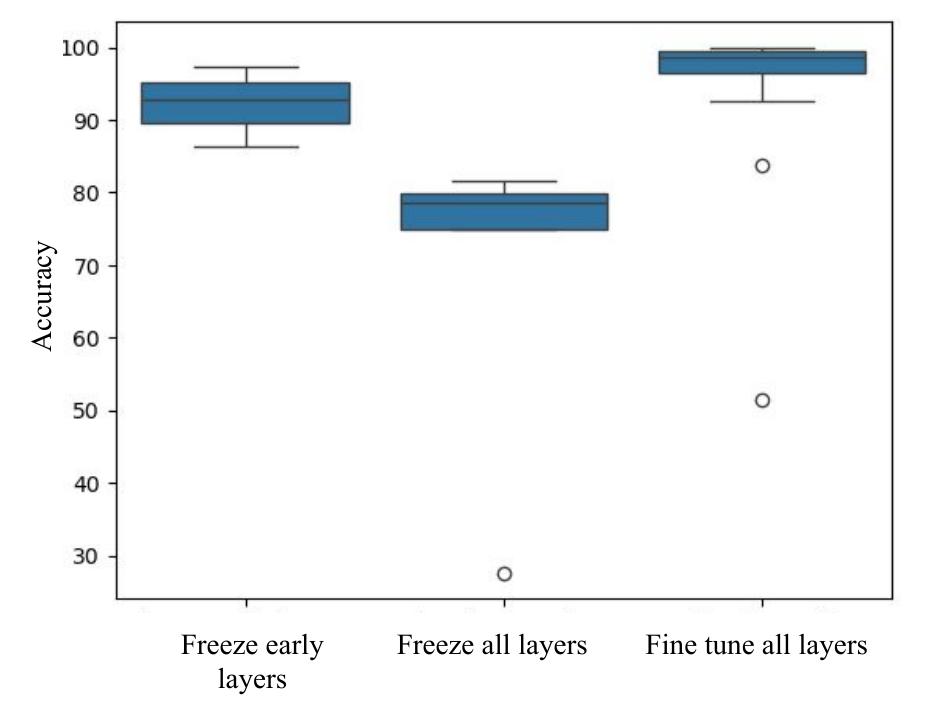} 
        \caption{Comparison of Adaptive Transfer learning: freezing early layers, Freeze all layers, and fine-tuning all layers.}
        \label{fig:params_mode_boxplot1}
    \end{minipage}
    \hfill
    \begin{minipage}{0.45\linewidth}
        \centering
        \includegraphics[width=\linewidth]{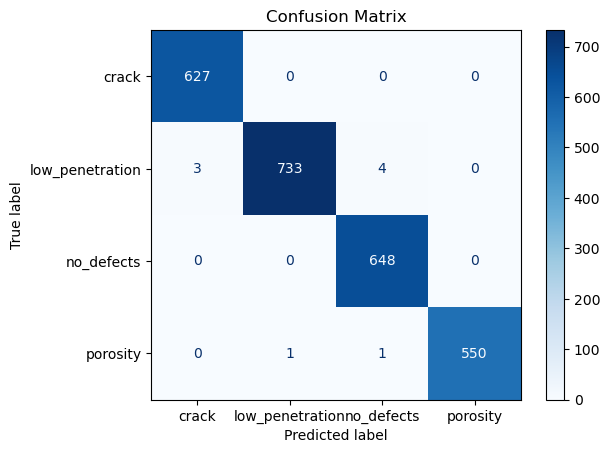} 
        \caption{Confusion matrix of the classifier, showcasing its performance across the four defect categories.}
        \label{fig:confusion_matrix}
    \end{minipage}
\end{figure}

The classifier, optimized using the best parameters identified by Adapt-Weld, was evaluated for its performance on the welding defect detection task. The model architecture selected is DenseNet121, fine-tuned in the "fine-tune all" mode, where all layers were updated during training. The optimizer used is AdamW with a learning rate of \(6.41 \times 10^{-5}\), chosen to ensure efficient convergence. A batch size of 16 was used for training to balance memory usage and model performance. The confusion matrix, presented in Figure~\ref{fig:confusion_matrix}, provides insights into the model's classification accuracy across the four defect categories: porosity (P), lack of penetration (LP), crack (C), and no defects.

\vspace{-0.2cm}

\vspace{-0.2cm}

\subsection{Qualitative XAI Results}
\label{xai_qual}

\begin{figure*}[htbp]
    \centering
    \includegraphics[width=1.2\textwidth]{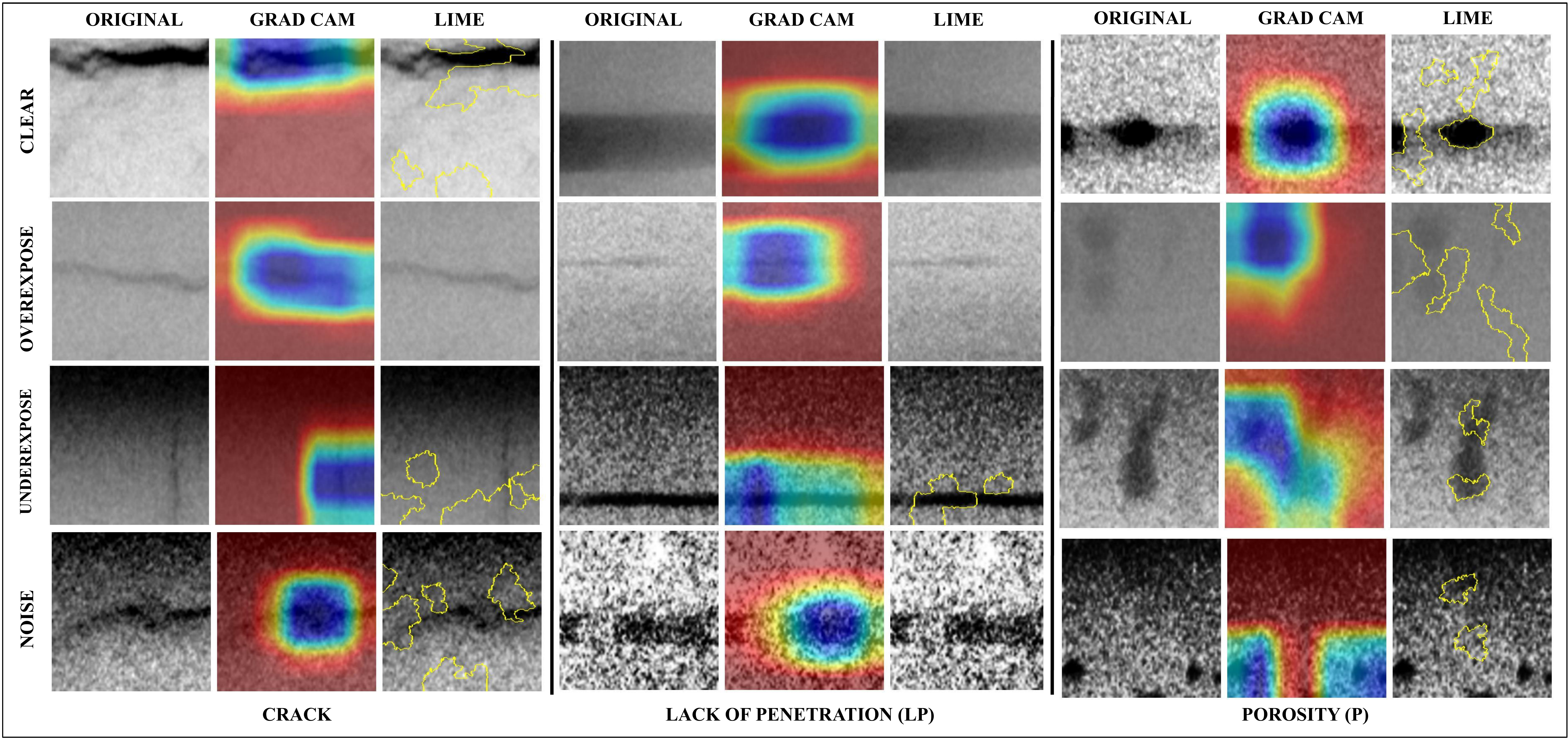}
    \caption{Explainability Analysis of Welding Defect Detection Using XAI Methods}
    \label{fig:xai}
\end{figure*}

The explainability analysis of welding defect detection using Grad-CAM and LIME demonstrates their effectiveness under different imaging conditions as represented in Figure~\ref{fig:xai}. It is observed that Grad-CAM provides a broad visualization of defect regions, highlighting the areas of interest across defect categories such as \textit{Crack}, \textit{Lack of Penetration (LP)}, and \textit{Porosity (P)}. The method effectively adapts to varying conditions, including clear, overexposed, underexposed, and noisy images. Whereas LIME complements Grad-CAM by offering localized, boundary-focused explanations. It highlights fine-grained regions, aiding in the detailed interpretation of the model's predictions. Grad-CAM is particularly reliable for identifying larger and more prominent defect regions, whereas LIME excels at pinpointing smaller, localized areas. However, under challenging conditions like noise and underexposure, Grad-CAM maintains robust defect localization, while LIME exhibits less consistent coverage but retains its ability to focus on specific regions.

Overall, the combination of these methods provides a comprehensive understanding of the model's decision-making process, ensuring reliable detection of welding defects even under challenging imaging scenarios.

\subsection{Quantitative Results of DDIA Framework Analysis}
\label{quant_ddia}
\begin{figure}[ht]
    \centering
    \includegraphics[width=\textwidth]{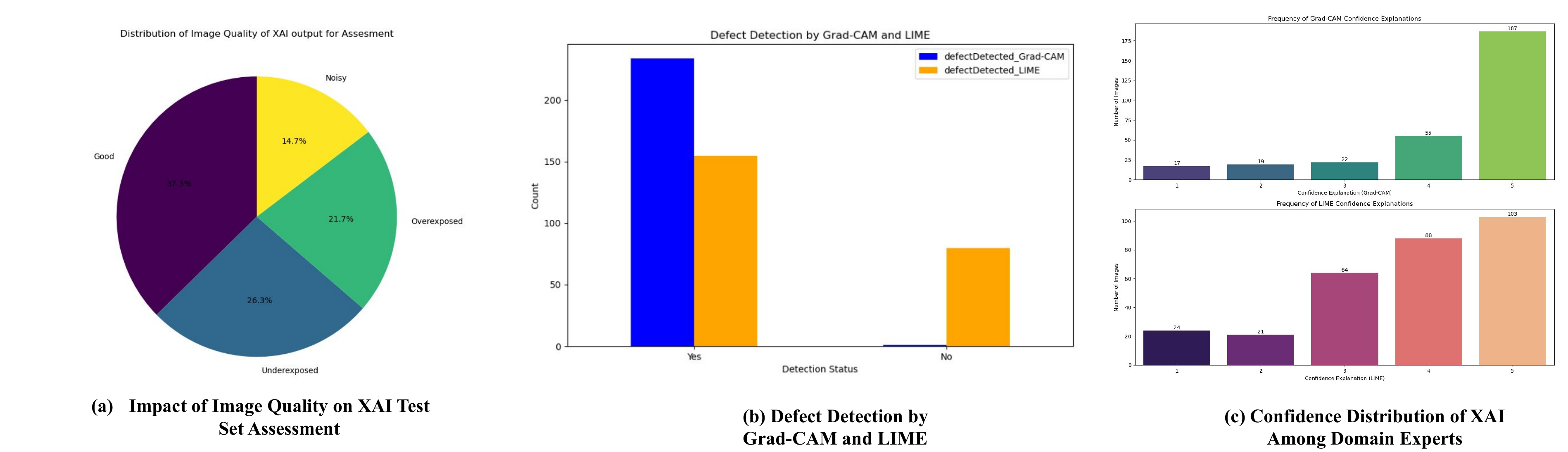}  
    \caption{Comparative analysis of Grad-CAM and LIME results: (a) Image Quality on XAI test set, (b) Defect Detection, and (c) Confidence of Domain Experts in XAI tool.}
    \label{fig:gradcam_lime_summary}
\end{figure}

The DDIA framework incorporates a \textbf{human-in-the-loop approach}, bridging the gap between domain expertise and AI applications to ensure safe deployment in maritime offshore structures.The comparative analysis in Figure~\ref{fig:gradcam_lime_summary} demonstrates: (a) \textbf{Impact of Image Quality on XAI Test Set Assessment}, where domain experts assess X-ray films categorized into noisy (14.7\%), overexposed (21.7\%), underexposed (26.3\%), and good quality films (37.3\%) to ensure that all critical scenarios are considered, enhancing the robustness of the evaluation;
 (b) \textbf{Defect Detection by Grad-CAM and LIME}, revealing that Grad-CAM provides clearer defect detection with fewer misdetections compared to LIME, making it a more reliable tool for identifying defects; and (c) \textbf{Confidence Distribution of XAI Among Domain Experts}, where confidence levels range from 1 (poor) to 5 (excellent). The distribution shows that Grad-CAM receives higher confidence scores, indicating that auditors find it more convincing and easier to use, while LIME scores slightly lower. This analysis highlights which XAI tool domain experts prefer and trust, reinforcing the framework's applicability in critical safety domains. Industrial experts can propose additional XAI techniques based on their specific needs, which will be evaluated according to company guidelines. This continuous feedback loop ensures the DDIA framework remains adaptable and incorporates cutting-edge XAI technologies,

\subsection{Quantitative Metrics for Interpretability in Defect Localization}
\label{quant_inter}
\vspace{-0.2cm}

\begin{figure}[ht]
    \centering
    \includegraphics[width=0.7\textwidth]{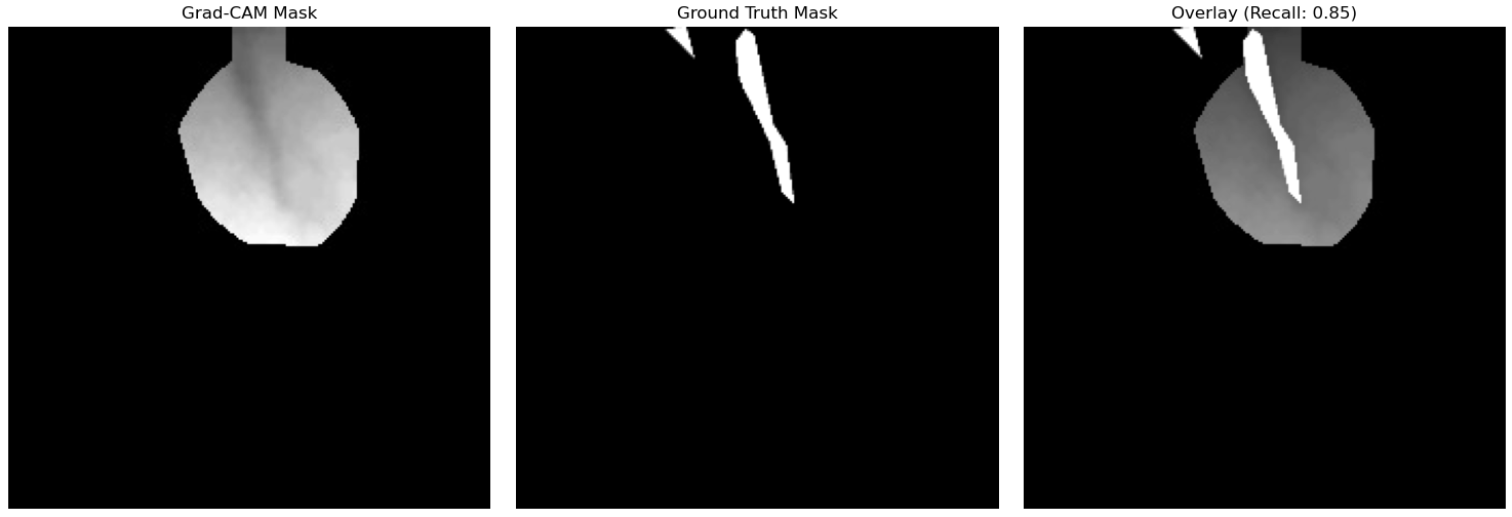}
    \caption{Recall-based evaluation showcasing the Grad-CAM mask, ground truth mask, and recall output for a sample image.}

    \label{fig:recall_example}
\end{figure}

To assess the interpretability and localization performance of Grad-CAM in identifying welding defects, we employed quantitative metrics based on recall. These metrics evaluate the overlap between the predicted defect regions (generated by Grad-CAM) and the ground truth annotations provided by domain experts. By quantifying this overlap, we aim to understand how effectively Grad-CAM highlights the defect regions in welding images.

An example of this evaluation is presented in Figure~\ref{fig:recall_example}, that visualizes the Grad-CAM mask, the ground truth mask, and the resulting recall output for a sample image. This analysis provides insights into the interpretability of Grad-CAM's predictions and its ability to focus on relevant defect regions. The average recall, calculated across the entire dataset, serves as a comprehensive metric to gauge the model’s performance in defect localization:
\vspace{-0.3cm}
\[
\text{Average Recall} = 0.7722
\]
This value highlights the ability of Grad-CAM to consistently localize defect regions, reinforcing its suitability as an XAI tool for welding defect analysis in safety-critical environments.

\vspace{-0.5cm}

\section*{Acknowledgment}
This work received the \textbf{Best Paper Award} at the International Conference on AI for the Oceans (ICAIO) 2025.  
The final version will be published in the \textit{Springer Lecture Notes in Networks and Systems (LNNS)}.

\section{Conclusion}
\vspace{-0.3cm}
This paper presents a framework for welding defect detection in offshore environments using adaptive AI models and Explainable AI (XAI). We introduce a novel framework "Adapt-WeldNet" that combines multiple pre-trained neural networks, optimizes hyperparameters and explores various training modes of transfer learning. XAI techniques, including Grad-CAM and LIME, enhance interpretability and transparency, which are crucial for high-risk applications like welding inspection. Additionally, we introduce the Defect Detection Interpretability Analysis (DDIA) framework, enabling direct interaction between AI models and domain experts, integrating a Human-in-the-loop approach to validate predictions and ensure safety. This DDIA framework supports Trustworthy AI in offshore structures by fostering transparency and accountability. A novel 'recall-based evaluation metric' is also proposed for the interpretability of XAI techniques for defect localization. Our results demonstrate the effectiveness of integrating adaptive AI with XAI to improve both accuracy and interpretability. This study provides a strong foundation for developing reliable, interpretable AI models for defect detection, enhancing operational efficiency, safety, and transparency in industrial environments. Future work will explore advanced XAI methods and extend the approach to other industrial sectors. 

\bibliographystyle{plain}       

\end{document}